\documentclass[letterpaper, 10 pt, conference]{ieeeconf}  

\IEEEoverridecommandlockouts                              
\usepackage{times}

\usepackage{multicol}

\usepackage{amsthm}
\usepackage{amsmath} 
\usepackage{amssymb}  
\usepackage{dsfont}
\usepackage{mathtools}
\usepackage{xcolor} 
\usepackage{ulem} 
\usepackage{multirow}
\usepackage{wrapfig}
\usepackage{bbm}
\usepackage{comment}

\usepackage{color}
\usepackage{latexsym}
\usepackage{multicol}

\usepackage{algorithm}
\usepackage{algpseudocode}

\usepackage[font=small,labelfont=bf]{caption}
\usepackage{lipsum}
\usepackage{subcaption}

\usepackage{url}

\renewcommand{\thefigure}{\arabic{figure} }

\newcommand{\Event}{\diamondsuit}

\newtheorem{definition}{Definition}
\newtheorem{problem}{Problem}

\newtheorem{proposition}{Proposition}

\title{\LARGE \bf
 Temporal Logic Guided Safe Reinforcement Learning Using Control Barrier Functions
}

\author{Xiao Li and Calin Belta
}

\begin{document}

\maketitle
\thispagestyle{empty}
\pagestyle{empty}

\begin{abstract}

Using reinforcement learning to learn control policies is a challenge when the task is complex with potentially long horizons. Ensuring adequate but safe exploration is also crucial for controlling physical systems. In this paper, we use temporal logic to facilitate specification and learning of complex tasks. We combine temporal logic with control Lyapunov functions to improve exploration. We incorporate control barrier functions to safeguard the exploration and deployment process. We develop a flexible and learnable system that allows users to specify task objectives and constraints in different forms and at various levels. The framework is also able to take advantage of known system dynamics and handle unknown environmental dynamics by integrating model-free learning with model-based planning.

\end{abstract}

\section{INTRODUCTION}

Our aim in this work is to address two challenges in current reinforcement learning (RL) methods - complex task specification and safe exploration. While much effort has been made in the development of sample efficient RL algorithms, the abilities to accurately specify the desired behavior and ensure safety during learning are paramount to the deployment of learning agents on physical systems.

Because reinforcement learning agents aim to maximize the expected return, they often exploit loopholes in the reward definition to gain an edge in achieving this goal and this results in undesired behavior. This phenomenon is often referred to as reward hacking \cite{Amodei2016} (some examples are provided in \cite{Lehman2018TheSC}). For tasks with higher complexity, an accurate definition of a reward function is challenging. For example, the authors of \cite{Vecerik2017LeveragingDF} shaped a reward function for the peg insertion task consisting of nested min/max functions that requires tuning to perform well. In this work, we use the robustness degree of temporal logic formulas as the reward function which is also constructed with nested min/max functions. However, the user only needs to specify the task as a logic formula and the transformation to the reward function is performed automatically. The robustness reward also has rigorous semantics such that a value greater than zero indicates completion of the task.

Safety is a critical aspect of reinforcement learning when physical systems that potentially operate around humans are involved. This issue is often addressed by constructing a similar environment in simulation, perform training in the simulation environment and transfer the learned policy to the real world \cite{James2018SimtoRealVS}\cite{Chebotar2018ClosingTS}. However, it is difficult for the dynamics of the simulation to exactly resemble that of reality and some fine-tuning is necessary for the sim-to-real transfer to succeed. Therefore, it is often an effective practice to incorporate a shielding controller both during learning and at deployment to ensure safety as well as to guide exploration \cite{Alshiekh2018SafeRL}\cite{Dalal2018SafeEI}.

In this paper, we propose a system that integrates temporal logic guided reinforcement learning with control barrier functions and control Lyapunov functions \cite{ames2014control}. Our main contributions are as follows:

\begin{itemize}
    \item we propose three ways to use temporal logic and its equivalent finite state automata (FSA) in our system - provide reward for the RL agent, perform goal selection for the control Lyapunov function for guided exploration and define safe sets for the control barrier functions.
    \item we extend the FSA augmented MDP framework \cite{li2018automata} to handle hard constraints and violation of specifications.
    \item we integrate the system into a quadratic program that can be solved efficiently.
    \item we demonstrate the applicability of our framework in a simulated continuous control task with safety constraints, known system dynamics and unknown environmental dynamics, and study the use and effectiveness of each component in the system.
\end{itemize}

\section{RELATED WORK}

Using temporal logic in an RL setting has been explored in the past. The authors of \cite{Giacomo2018ReinforcementLF} and \cite{camacho2017decision} use temporal logic and automata to solve the non-Markovian reward decision process (NMRD). In \cite{aksaray2016q}, the authors take advantage of the robustness degree of signal temporal logic to guide the learning process in discrete MDPs. In \cite{icarte2018using}, the authors propose a reward machine, which in effect is an FSA. However, the user is required to manually design the reward machine. In \cite{Sadigh2014} and \cite{Littman2017EnvironmentIndependentTS}, linear temporal logic (LTL) and geometric linear temporal logic (GLTL) are combined with MDP to learn policies in discrete environments.

Safe exploration and deployment is critical for applying learning agents on physical systems. The authors of \cite{Dalal2018SafeEI} and \cite{Achiam2017ConstrainedPO} approach the problem in the context of constrained policy optimization and showed minimum violation of safety constraints during learning. The authors of \cite{Ohnishi2018BarrierCertifiedAR} and \cite{endtoendcbf} utilize control barrier functions to safeguard exploration. Temporal logic is used in conjunction with an abstraction of the system dynamics to shield the learning process from unsafe actions in \cite{Alshiekh2018SafeRL}. The Authors of \cite{Garca2015ACS} provides a comprehensive survey on safe reinforcement learning. We incorporate hard constraints in a TL specification with control barriers functions such that these constraints can be strictly enforced (in addition to other user specified constraints) during learning. Compared to previous work, our method is relatively easy to implement, able to handle continuous state and action spaces, nonlinear constraints and dynamics, and can be executed efficiently.

\section{PRELIMINARIES}

\subsection{Reinforcement Learning}

We start with the definition of a Markov Decision Process.

\begin{definition}\label{def2}
An MDP is defined as a tuple $\mathcal{M} = \langle S,A,p(\cdot|\cdot,\cdot),r(\cdot,\cdot, \cdot)\rangle$, where $S\subseteq {\rm I\!R}^n$ is the state space ; $A \subseteq {\rm I\!R}^m$ is the action space ($S$ and $A$ can also be discrete sets); $p: S \times A \times S \to [0,1]$ is the transition function with $p(s^{\prime}|s,a)$ being the conditional probability density of taking action $a \in A$ at state $s \in S$ and ending up in state $s^{\prime} \in S$; $r: S \times A \times S \to {\rm I\!R}$ is the reward function with $r(s,a,s^\prime)$ being the reward obtained by executing action $a$ at state $s$ and transitioning to $s^\prime$.
\end{definition}

We define a task to be the process of finding the optimal policy $\pi^\star: S \to A$ (or $\pi^\star: S \times A \to [0,1]$ for stochastic policies) that maximizes the expected return, i.e.

\begin{equation}\label{eq2B1}
\pi^\star = \underset{\pi}{\arg\max}\mathbb{E}^\pi[\sum_{t=0}^{T-1} r(s_t, a_t, s_{t+1})],
\end{equation}
The horizon of a task (denoted $T$) is defined as the maximum allowable time-steps of each execution of $\pi$ and hence the maximum length of a trajectory.
In Equation~\eqref{eq2B1}, $\mathbb{E}^\pi[\cdot]$ is the expectation following $\pi$. The state-action value function is defined as

\begin{equation}\label{eq2B2}
Q^\pi(s,a) = \mathbb{E}^\pi[\sum_{t=0}^{T-1} r(s_t, a_t, s_{t+1})|s_0=s, a_0=a]
\end{equation}
\noindent i.e. it is the expected return of choosing action $a$ at state $s$ and following $\pi$ onwards. For off-policy actor critic methods such as deep deterministic policy gradient~\cite{Lillicrap2015ContinuousCW}, $Q^\pi$ is used to evaluate the quality of policy $\pi$. Parameterized $Q^\pi_{\theta_Q}$ and $\pi_{\theta_\pi}$ ($\theta_Q$ and $\theta_\pi$ are learnable parameters) are optimized alternately to obtain $\pi^\star_{\theta_\pi}$.

\subsection{scTLTL And Finite State Automata}
We consider tasks specified with \textit{syntactically co-safe Truncated Linear Temporal Logic} (scTLTL) which is a restricted version of truncated linear temporal logic(TLTL)~\cite{li2016reinforcement}. In particular, we restrict from using the $\Box$ (always) operator. By doing so, we can establish a correspondence between an scTLTL formula with a FSA.

Due to space constraints, we do not provide the complete set of definitions for the Boolean and quantitative semantics of scTLTL (refer to \cite{li2016reinforcement}). Examples of scTLTL include $\diamondsuit (\phi_a \wedge \diamondsuit \phi_b)$ which entails that \textit{eventually} $\phi_a$ and then \textit{eventually} $\phi_b$ become true (sequencing). Another example $(\phi_a \Rightarrow \diamondsuit \phi_b) \,\, \mathcal{U}\,\, \phi_c $ means \textit{until} $\phi_c$ becomes true, $\phi_a$ is true \textit{implies} that \textit{eventually} $\phi_b$ is true.

We denote $s_t \in S$ to be the MDP state at time $t$, and $s_{t:t+k}$ to be a sequence of states (state trajectory) from time $t$ to $t+k$, i.e., $s_{t:t+k}=s_ts_{t+1}...s_{t+k}$. scTLTL provides a set of real-valued functions that quantify the degree of satisfaction of a given $s_{0:T}$ with respective to a formula $\phi$. This measure is also referred to as robustness degree or simply robustness ($\rho(s_{0:T}, \phi)$ maps a state trajectory and a formula to a real number). For example, $\rho(s_{0:3}, \Event (s < 4)) = \max(4-s_0, 4-s_1, 4-s_2)$. Here, if $4-s_t>0$, then $s_t < 4$ is satisfied. Because $\Event (s < 4)$ requires $s < 4$ to be true at least once in the trajectory, hence we take the max over the time horizon. In general, a robustness of greater than zero implies that $s_{t:t+k}$ satisfies $\phi$ and vice versa.

\begin{definition}\label{def1}
An FSA corresponding to a scTLTL formula $\phi$. is defined as a tuple  $\mathcal{A}_\phi=\langle \mathbb{Q}_\phi, \Psi_\phi, q_{\phi, 0}, p_\phi(\cdot | \cdot), \mathcal{F}_\phi \rangle $, where $\mathbb{Q}_\phi$ is a set of automaton states; $\Psi_\phi$ is the input alphabet (a set of first order logic formulas); $q_{\phi, 0} \in \mathbb{Q}_\phi$ is the initial state; $p_\phi:\mathbb{Q}_\phi \times \mathbb{Q}_\phi \rightarrow [0,1]$ is a conditional probability defined as

\begin{equation}\label{eq2A2}
\begin{split}
p_\phi(q_{\phi, j} | q_{\phi, i}) &= \begin{cases}
1 & \psi_{q_{\phi, i}, q_{\phi, j}} \textrm{ is true} \\
 0 & \text{otherwise}.
 \end{cases} \\
 \quad\quad\quad\qquad or  \\
 p_\phi(q_{\phi, j} | q_{\phi, i}, s) &= \begin{cases}
1 & \rho(s,\psi_{q_{\phi, i}, q_{\phi, j}})>0\\
 0 & \text{otherwise}.
 \end{cases}
\end{split}
\end{equation}

\noindent $\mathcal{F}_\phi$ is a set of final automaton states.

\end{definition}

Here $q_{\phi, i}$ is the $i^{th}$ automaton state of $\mathcal{A}_\phi$. $\psi_{q_{\phi, i}, q_{\phi, j}} \in \Psi_\phi$ is the predicate guarding the transition from $q_{\phi, i}$ to $q_{\phi, j}$. Because $\psi_{q_{\phi, i}, q_{\phi, j}}$ is a predicate without temporal operators, the robustness $\rho(s_{t:t+k}, \psi_{q_{\phi, i}, q_{\phi, j}})$ is only evaluated at $s_t$. Therefore, we use the shorthand $\rho(s_{t}, \psi_{q_{\phi, i}, q_{\phi, j}}) = \rho(s_{t:t+k}, \psi_{q_{\phi, i}, q_{\phi, j}})$. The FSA corresponding to a scTLTL formula can be generated automatically with available packages like Lomap \cite{lomap} (refer to \cite{rozier2013explicit} for details on the generation procedure).

\subsection{FSA Augmented MDP}

The FSA augmented MDP $\mathcal{M}_\phi$ \cite{li2018automata} establishes a connection between the TL specification and the standard reinforcement learning problem. A policy learned using $\mathcal{M}_\phi$ has implicit knowledge of the FSA through the automaton state $q_{\phi} \in \mathcal{Q}_\phi$.

\begin{definition}\label{def3}
An FSA augmented MDP corresponding to a scTLTL formula $\phi$ (constructed from FSA $\langle \mathbb{Q}_\phi, \Psi_\phi, q_{0}, p_\phi(\cdot | \cdot), \mathcal{F}_\phi \rangle$ and MDP $\langle S,A,p(\cdot|\cdot,\cdot),r(\cdot,\cdot, \cdot)\rangle$) is defined as $\mathcal{M} _\phi= \langle \tilde{S}, A, \tilde{p}(\cdot|\cdot,\cdot),\tilde{r}(\cdot, \cdot),  \mathcal{F}_\phi \rangle$, where $\tilde{S} \subseteq S \times \mathbb{Q}_{\phi}$, $\tilde{p}(\tilde{s}'|\tilde{s},a)$ is the probability of transitioning to $\tilde{s}^\prime$ given $\tilde{s}$ and $a$,

 \begin{equation}\label{eq3A2}
 \tilde{p}(\tilde{s}'|\tilde{s},a) = p\big((s', q')|(s,q), a\big)
 = \begin{cases}
 p(s'|s,a) & p_\phi(q'|q,s) =1 \\
 0 & \text{otherwise}.
 \end{cases}
 \end{equation}

\noindent $p_\phi$ is defined in Equation~\eqref{eq2A2}. $\tilde{r}: \tilde{S} \times \tilde{S} \to {\rm I\!R}$ is the FSA augmented reward function, defined by


\begin{equation}\label{eq3A3}
\tilde{r}(\tilde{s},\tilde{s}') = \rho(s',D_{\phi}^{q}),
\end{equation}
\noindent where $D_{\phi}^{q}=\bigvee_{q^\prime \in\Omega_{q}} \psi_{q, q^\prime}$ represents the disjunction of all predicates guarding the outgoing transitions that originate from $q$ ($\Omega_{q}$ is the set of automata states that are connected with $q$ through outgoing edges).
\end{definition}

\noindent Note that in Equation~\eqref{eq3A3}, $\tilde{r}$ is calculated from $(s^\prime, q)$. It is a measure of the progress of satisfying $D_{\phi}^{q}$ by taking action $a$ in state $s$ (encapsulated by $s^\prime$).

Intuitively, the reward function in Equation~\eqref{eq3A3} encourages the system to exit the current automaton state and move on to the next, and by doing so eventually reach the final state $q_f$ which satisfies the TL specification (property of FSA).

\subsection{Zeroing Control Barrier Functions (CBF) And Control Lyapunov Functions (CLF)}

Define an affine control system as

\begin{equation}\label{eq3d1}
    \Dot{s} = f(s) + g(s)a
\end{equation}

\noindent where $f: S \rightarrow S$ and $g: A \rightarrow S$ are locally Lipschitz continuous, $s \in S \subseteq  {\rm I\!R}^n$ is the MDP state, $a \in A \subseteq  {\rm I\!R}^m$ is the control. Here we used the same notation for state and action as the MDP in Definition \ref{def2}. As will become clear in the next section, we embed the control system as part of the MDP transition dynamics. Following \cite{xu2015robustness} and \cite{ames2014control}, we provide the definition of the zeroing control barrier function and the control Lyapunov function.

\begin{definition}\label{def3d}
 Given a set $\mathcal{C}=\{s \in S: h(s) \geq 0\}$ for a continuously differentiable function $h: S \rightarrow {\rm I\!R}$, the function $h$ is a zeroing control barrier function defined on set $\mathcal{D}$ with $\mathcal{C} \subseteq \mathcal{D} \subset {\rm I\!R}$, if there exists an extended class $\mathcal{K}$ function $\alpha$ such that

 \begin{equation}\label{eq3d2}
     \underset{s}{sup}\big(\frac{\partial h(s)}{\partial t} + \alpha h(s) \geq 0 \big).
 \end{equation}
\end{definition}

\begin{definition}
 A continuously differentiable function $V:S \rightarrow {\rm I\!R}$ is an exponentially stabilizing control Lyapunov function \cite{ames2014control} if there exists positive constants $c_1, c_2, c_3 > 0$ such that

 \begin{equation}\label{eq3d3}
     \begin{split}
         & c_1||s||^2 \leq V(s) \leq c_2 ||s||^2 \\
         & \underset{s}{\inf} (\frac{\partial V(s)}{\partial t} + c_3 V(s) \leq 0).
     \end{split}
 \end{equation}
\end{definition}

\begin{proposition}
A controller that meets the objectives specified by the CLF and renders the safe set $\mathcal{C}$ forward invariant (satisfies the condition in Equation \eqref{eq3d2}) can be found by solving the quadratic program (CLF-CBF-QP)

\begin{equation}
    \begin{split}
       &a^\star(s) = \underset{a \in A, \delta}{\arg\min} \,\, a^Ta + K\delta\\
       s.t.\,\, & \frac{\partial h(s)}{\partial s}f(s) + \frac{\partial h(s)}{\partial s}g(s)a+ \alpha h(s) \geq 0\\
       & \frac{\partial V(s)}{\partial s}f(s) + \frac{\partial V(s)}{\partial s}g(s)a+ c_3 V(s) \leq \delta,
    \end{split}
\end{equation}

\noindent where $\delta$ is a relaxation variable that grows when there is a conflict between the CBF and CLF constraints. $K$ is chosen to be a large positive constant ($\sim 1e^{10}$).
\end{proposition}


\section{PROBLEM FORMULATION AND APPROACH}

\begin{problem}\label{prob1}
Given an MDP  $\mathcal{M} = \langle S,A,p(\cdot|\cdot,\cdot),r(\cdot,\cdot, \cdot)\rangle$ where $S \subseteq {\rm I\!R}^n$ is the state space; $A \subseteq {\rm I\!R}^m$ is the action space; $p=\{p_{k}, p_{u}\}$ consists of known system dynamics in the form $p_k: \Dot{s}=f(s)+g(s)a$ where $s \in S, \, a \in A $ and unknown environmental dynamics $p_u$; $r: S \times A \times S \rightarrow {\rm I\!R}$ is the reward function, a safe set $\mathcal{C}=\{s \in S: h(s) \geq 0\}$ and a scTLTL formula $\phi$ over state predicates, find a policy $\pi^\star_\phi$ such that:


\begin{equation}\label{eq3A1}
\begin{split}
& \pi^\star_\phi = \underset{\pi_\phi}{\arg \max}\mathbb{E}^{\pi_\phi}[ \sum_{t=0}^T \gamma^t (\tilde{r} + wr)] \\
s.t.\,\, & \underset{t \in [0, T]}{\min} h(s_t) > 0,
\end{split}
\end{equation}

\noindent where $\tilde{r}$ is the FSA augmented MDP reward in Equation \eqref{eq3A3}, $w$ is a weighting parameter, $T$ is the time horizon and $0<\gamma \leq 1$ is the discount factor, $h$ is the control barrier function in Definition \ref{def3d}. It is required that the system trajectory always stays within the safe set defined by $h$ and $\phi$ is not violated at all times during learning.

\end{problem}

\noindent Equation~\eqref{eq3A1} requires $\pi^\star_\phi$ to optimize the expectation of a weighted return that combines satisfying $\phi$ (maximizing $\tilde{r}$) and maximizing the MDP reward $r$ while ensuring safety during learning. Here we distinguish between not satisfying $\phi$ and violating $\phi$. Violation means that a state is entered such that $\phi$ can never be satisfied afterwards.

\begin{figure}[tbh]
\begin{center}
\includegraphics[width=.8\linewidth]{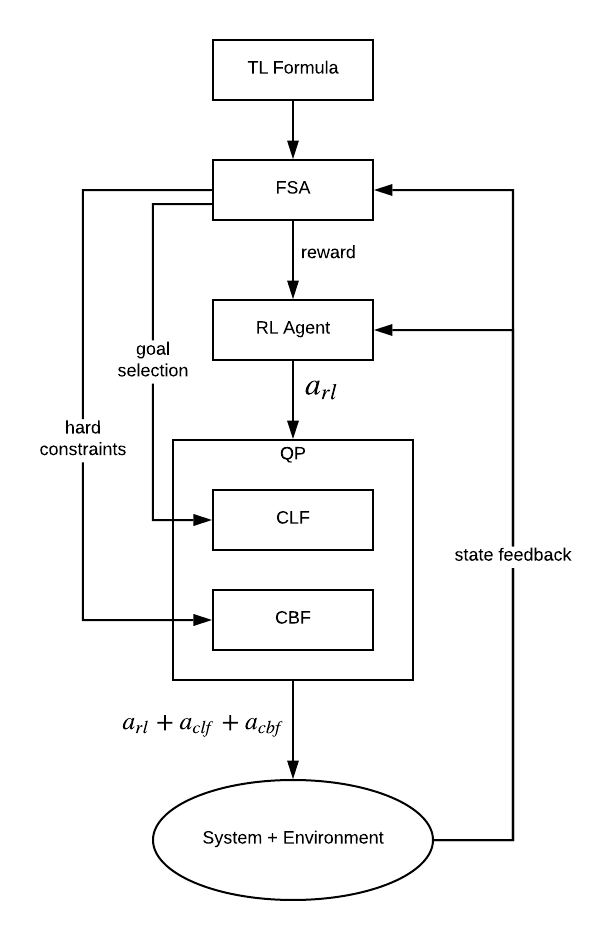}
\caption{Flow diagram for TL guided safe reinforcement learning}\label{fig:4.1}
\end{center}
\end{figure}

To solve Problem \ref{prob1}, we first translate $\phi$ to its equivalent FSA and construct a FSA augmented MDP using Definition \ref{def3}. The FSA is used in three ways - provide reward for the RL agent, provide goal states for the CLF to guide exploration and provide the safe set for the CBF. The actions from the RL agent are then integrated with the CLF-CBF-QP to solve for the final safe action that's transmitted to the system and environment. A diagram of our workflow is provided in Figure \ref{fig:4.1}.

\section{TL GUIDED SAFE REINFORCEMENT LEARNING}

\begin{figure}[tbh]
\begin{center}
\includegraphics[width=0.9\linewidth]{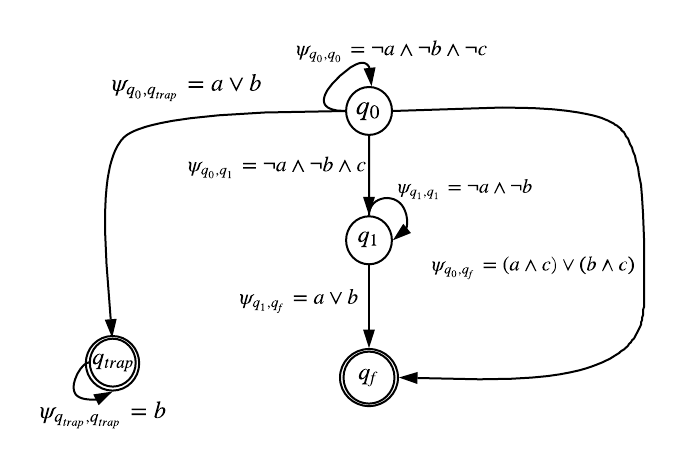}
\caption{FSA for formala $(\diamondsuit a \vee \diamondsuit b) \wedge \diamondsuit c \wedge (\neg a \wedge \neg b \,\, \mathcal{U} \,\, c)$}\label{fig:5.1}
\end{center}
\end{figure}

In this section, we discuss in detail the components of Figure \ref{fig:4.1}. We will use a running example for explanation purposes. We define three circular regions $a$, $b$, $c$ each of form $|s - s_i| < th, \, i \in \{a,b,c\}$ where $s_i$ is the center coordinate of the region, $th$ is the radius. A specification $\phi=(\diamondsuit a \vee \diamondsuit b) \wedge \diamondsuit c \wedge ((\neg a \wedge \neg b) \,\, \mathcal{U} \,\, c)$ entails that regions $a$ \textit{or} $b$ \textit{and} $c$ are to be \textit{eventually} visited \textit{and} $a$ \textit{and} $b$ should \textit{not} be visited before $c$. The FSA for $\phi$ is shown in Figure \ref{fig:5.1}.

One drawback of the FSA definition (Definition \ref{def1}) is that it does not explicitly handle spec violation. This problem can be solved by adding an additional set of trap states $Q_{trap}$ such that if $q_{trap} \in Q_{trap}$, then no path exists on the FSA that connects $q_{trap}$ to $q_f$ (i.e. entering $q_{trap}$ violates the spec). This can be seen in Figure \ref{fig:5.1}when $b$ is visited before $a$.

\subsection{FSA as RL Reward}

Here we extend the reward definition in Equation \eqref{eq3A3} for the FSA augmented MDP to handle the trap state. The modified reward is as follows:

\begin{equation}\label{eq5A1}
    \tilde{r} = \tilde{r}(\tilde{s},\tilde{s}') = \min \big(\rho(s',D_{\phi}^{q}), c_r\rho(s^\prime, \neg \psi_{q, q_{trap}}) \big),
\end{equation}

\noindent where $D_{\phi}^{q}=\underset{q^\prime \in\Omega_{q}, q^\prime \neq q_{trap}}{\bigvee} \psi_{q, q^\prime}$. As an example, in Figure \ref{fig:5.1} $D_{\phi}^{q_0} = \psi_{q_0, q_1} \vee \psi_{q_0, q_f}$.

$\rho(s',D_{\phi}^{q})$ in Equation \eqref{eq5A1} promotes exiting the current FSA state and enter the next non-trap state. $\rho(s^\prime, \neg \psi_{q, q_{trap}})$ penalizes entering the trap state. the min() function penalizes the more violating term with lower robustness. $c_r>0$ is a weighting coefficient. It is important to note that $\tilde{r}$ in Equation \eqref{eq5A1} does not prevent the RL agent from violating $\phi$ during learning. In fact, the agent needs to explore $q_{trap}$ in order to learn and hence the self-loop. During testing, the user can choose to terminate once the a trap state is entered

\subsection{FSA For CLF Goal Selection}

The intuition behind goal selection using the FSA is to find the outgoing edge (leading to a non-trap state) that is easiest for the agent to activate at its current state. Then select the MDP state that maximizes the robustness of the guarding predicate of that edge as the goal. This can be formalized by

\begin{equation}\label{eq5B1}
\begin{split}
    & s_g(q) = \underset{s \in S}{\arg\max} \,\, \rho(s, \psi_{q, q^\prime_{\rho_{max}}}), \textrm{ where}\\
    & q^\prime_{\rho_{max}}(s,q) = \underset{q^\prime \in \Omega_q, q^\prime \neq q_{trap}}{\arg\max} \rho(s, \psi_{q,q^\prime}),
\end{split}
\end{equation}

\noindent where $\Omega_{q}$ is the set of automata states that are connected with $q$ through outgoing edges. Finding $s_g$ using Equation \eqref{eq5B1} requires solving two optimization problems. The second one is relatively easy since there are only a finite number of outgoing edges to choose from. The first problem can be difficult if the state space is large. One useful heuristic is that if the predicates are defined as regions in the state space, the state with max robustness is the center of that region or at the center of intersection of multiple regions. For the example in Figure~\ref{fig:5.1}, $s_g(s,q_1)$ would be the center of region $a$ or $b$ whichever is closer to state $s$.

After finding $s_g$, a simple CLF can be defined as

\begin{equation}\label{eq5B2}
    V_\phi(s,q) = (s-s_g)^T(s-s_g).
\end{equation}

\noindent Other CLFs that meet the criteria in Equation \eqref{eq3d3} can also be used. For simplicity, the above CLF will be adopted in our case studies.

\subsection{FSA for CBF Constraints}

We exploit the fact that the guarding predicates $\psi$ in an FSA are propositional logic formulas of inequality constraints over states. They can be naturally interpreted as safe sets and incorporated into the CBF. At state $(s,q)$, if there is a trap state $q_{trap}$ connected to $q$, we would like to avoid activating $\psi_{q, q_trap}$. That is we need to ensure that $\psi_{q, q_{trap}}$ is always false or $\neg \psi_{q, q_{trap}}$ is always true. We first restructure $\neg \psi_{q, q_{trap}}$ to its conjunctive normal form (CNF) i.e. $\neg \psi_{q, q_{trap}} = \bigwedge_i (\bigvee_{j} \psi_{ij})$ \cite{jackson2004clause} - the conjunction of a set of disjunctive formulas. An example of a CNF is $\bigwedge_i (\bigvee_{j} \psi_{ij}) = (a) \wedge (b \vee \neg c)$ where $\phi_{00} = a$, $\phi_{10} = b$, $\phi_{11} = \neg c$.

We define the FSA-based safe set at state $(s,q)$ with $\neg \psi_{q, q_{trap}} = \bigwedge_i (\bigvee_{j} \psi_{ij})$ as

\begin{equation}\label{eq5C1}
\begin{split}
    &\mathcal{C}_\phi(s,q) = \{h^0_\phi(s,q),\, ...\, , h^i_\phi(s,q)\}. \textrm{ where}\\
    & h^i_\phi(s,q) = \psi_{ij_{\rho_{min}}}, \,\, j_{\rho_{min}} = \underset{j}{\arg\min} \rho(s, \psi_{ij})
\end{split}
\end{equation}

\noindent Here we assume each predicate $\psi_{ij}$ is written in the form $f(s)>0$. The goal of the CBF is to keep the agent from entering the trap state during learning

For the example in Figure \ref{fig:5.1}, at state $(s,q_0)$, $\psi_{q_0,q_{trap}} = a \vee b$, $\neg \psi_{q_0,q_{trap}} = \neg a \wedge \neg b$. Therefore, $\mathcal{C}_\phi(s,q_0) = \{\neg a, \neg b\} = \{|s - s_a| > th, |s-s_b| > th\}$.

\subsection{TL-CLF-CBF QP}

We modify the known system dynamics $p_k$ to

\begin{equation}\label{eq5D1}
\begin{split}
    & f(s) + g(s)(a_{rl}(s)+a_{cbf}+a_{clf}) \\
    &= \underbrace{f(s)+g(s)a_{rl}(s)}_{\hat{f}(s)} + g(xs)(a_{cbf}+a_{clf}),
\end{split}
\end{equation}

\noindent $a_{rl}$ is the action provided by the RL agent. $a_{cbf}$ and $a_{clf}$ can be found by solving the following quadratic program

\begin{equation}\label{eq5D2}
    \begin{split}
       \hat{a}^\star(s, q) &= \underset{\hat{a}, \delta}{\arg\min} \,\, \hat{a}^T\hat{a} + K\delta\\
       s.t.\,\, & \frac{\partial h(s)}{\partial s}\hat{f}(s) + \frac{\partial h(s)}{\partial s}g(s)\hat{a}+ \alpha h(s) \geq 0\\
       &\frac{\partial h^0_\phi(s,q)}{\partial s}\hat{f}(s) + \frac{\partial h^0_\phi(s,q)}{\partial s}g(s)\hat{a}+ \alpha h^0_\phi(s,q) \geq 0\\
       & \,\,\,\,\,\,\,\,\,\,\,\,\,\,\,\,\,\,\,\,\,\,\,\,\,\,\,\,\,\,\,\,\,\,\vdots \\
       & \frac{\partial h^n_\phi(s,q)}{\partial s}\hat{f}(s) + \frac{\partial h^n_\phi(s,q)}{\partial s}g(s)\hat{a}+ \alpha h^n_\phi(s,q) \geq 0\\
       & \frac{\partial V_\phi(s,q)}{\partial s}\hat{f}(s) + \frac{\partial V(s,q)}{\partial s}g(s)\hat{a}+ c_3 V(s,q) \leq \delta\\
       & a_{cbf}^{min} \leq a_{cbf} \leq a_{cbf}^{max}\\
       & a_{clf}^{min} \leq a_{clf} \leq a_{clf}^{max}
    \end{split}
\end{equation}

\noindent where $\hat{a} = [a_{cbf}, a_{clf}]$, $h(s)$ captures extra safety constraints defined by the user, $h^i_\phi \in \mathcal{C}_\phi, \, i \in \{1, ..., n\}$ are TL spec related constraints constructed with Equation \eqref{eq5C1}, $V_\phi$ is constructed with Equation \eqref{eq5B2}. A summary of our algorithm is provided in Algorithm \ref{alg:1}.

\begin{algorithm}
\caption{TL Guided Safe RL}
\label{alg:1}
\begin{algorithmic}[1]
\State \textbf{Inputs}: scTLTL task specification $\phi$, MDP $\mathcal{M}$, episode horizon $T$, RL agent $\mathcal{RL}$, parameterized policy $\pi_\theta$, action bounds $b = [a_{cbf}^{max},a_{cbf}^{min},a_{clf}^{max},a_{clf}^{min}]$, QP parameters $c_r$, $K$, $\alpha$, $c_3$, reward weighting parameter $w$
\State Construct the FSA augmented MDP $\mathcal{M}_\phi$ \Comment{using Definition 3}
\State Initialize policy $\theta \leftarrow \theta_0$, empty episode buffer $\mathcal{B} \leftarrow []$
\For {i = 0 \textrm{ to number of training iterations}}
\State Reset environment
\State Receive initial observation $s_0$
\For {$t = 0 \textrm{ to }T$}
\State $a_{rl}^t = \pi_{\theta_i}(s_t)$
\State Construct Lyapunov function $V_\phi$ and safe set $\mathcal{C}_\phi$ \Comment{Equation \eqref{eq5C1}, \eqref{eq5B2}}
\State $a_{cbf}^t$, $a_{clf}^t = QP(s_t, q_t, a_{rl}^t,\mathcal{C}_\phi, V_\phi)$ \Comment{Equation \eqref{eq5D2}}
\State $s_{t+1} \leftarrow step(s_t, q_t, a_{rl}+a_{cbf}+a_{clf})$ \Comment{
step the environment to get next observation}
\State Construct reward $\tilde{r}+wr$ \Comment{$\tilde{r}$ from Equation \eqref{eq5A1}}
\State $\mathcal{B}.append\big((s_t, q_t a_rl, s_{t+1}, \tilde{r}+wr)\big)$
\EndFor
\State $\theta_i \leftarrow \mathcal{RL}(\theta_{i-1}, \mathcal{B})$ \Comment{update the policy with chosen RL agent}
\EndFor\\
\Return optimal policy parameters $\theta^\star$
\end{algorithmic}
\end{algorithm}

This framework provides much flexibility in defining a task. It incorporates prior knowledge about the task domain in the form of temporal logic specifications as well as regular reward functions. The user can also specify hard and soft constraints (encapsulated in the CBF and CLF). The system also combines model-based planning and model-free learning so that knowledge about the system dynamics can be taken advantage of while unknown dynamics can also be handled.

\section{CASE STUDY}

\subsection{Experiment Setup}

\begin{figure}[tbh]
\begin{center}
\includegraphics[width=.8\linewidth]{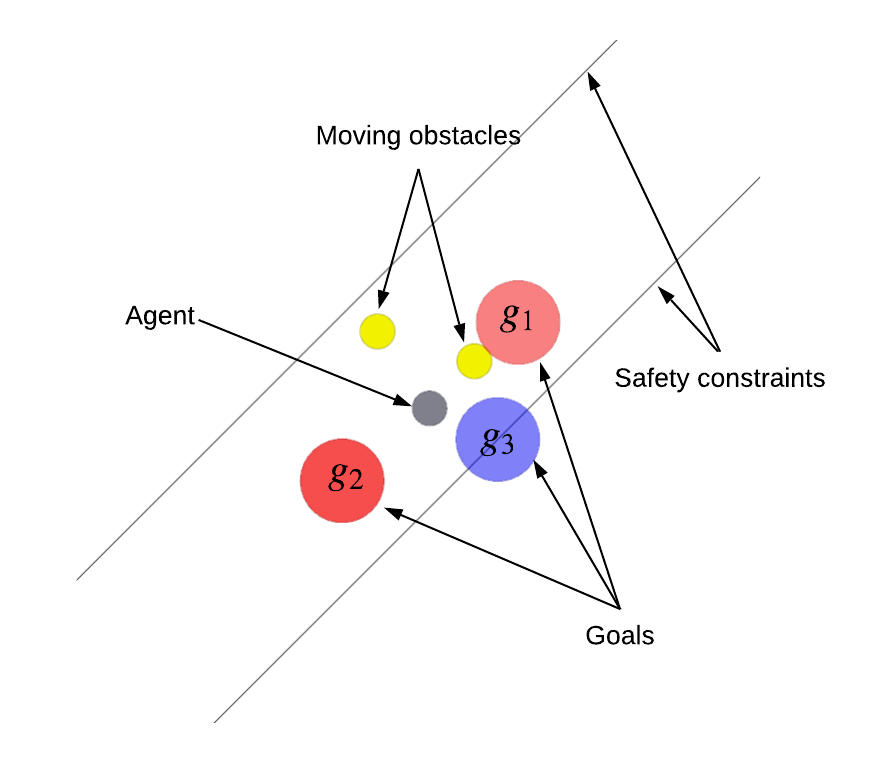}
\caption{Simulation environment setup.}\label{fig:6.1.1}
\end{center}
\end{figure}

Figure \ref{fig:6.1.1}shows our simulation environment. It consists of one agent with known dynamics, safety constraints in the form of two straight lines forming a channel that the agent has to stay within, three circular goal regions whose positions are kept fixed in a episode but are randomized between episodes, and two obstacles that move in the vicinity of the channel and whose dynamics are unknown.

\subsection{Agent Dynamics}

The MDP state of the agent consists of its position and orientation $[x,y,\theta]$ as well as the position of all goals and moving obstacles (13 continuous dimensions). Combined with 1 discrete dimension for the FSA state (from construction of the FSA augmented MDP), the total state space is 14 dimensional. The agent's controls are the forward velocity $a_v$ and steering speed $a_\theta$.

The agent moves according to the unicycle model
\begin{equation}\label{eq6A1}
    \begin{bmatrix}
    \Dot{x}\\
    \Dot{y}\\
    \Dot{\theta}
    \end{bmatrix} = \begin{bmatrix} \cos(\theta) & 0 \\ \sin(\theta) & 0 \\ 0 & 1 \end{bmatrix} \begin{bmatrix} a_v \\ a_\theta \end{bmatrix}
\end{equation}

\noindent This underactuated system is uncontrollable. As in \cite{belta2005discrete}\cite{desai1998controlling}, we use a reference point different from the robot center with coordinates $[\epsilon, 0],\, \epsilon>0$ in the robot frame. It is shown that the relationship below holds

\begin{equation}\label{eq6A2}
    \begin{bmatrix} \Dot{x}_\epsilon \\ \Dot{y}_\epsilon \end{bmatrix} = \begin{bmatrix} \cos(\theta) & -\sin(\theta) \\ \sin(\theta) & \cos(\theta) \end{bmatrix}\begin{bmatrix} 1 & 0 \\ 0 & \epsilon \end{bmatrix}\begin{bmatrix} a_v \\ a_\theta \end{bmatrix},
\end{equation}

\noindent where $[x_\epsilon, y_\epsilon]$ are the coordinates of the reference point (orientation is the same as the center). With $a_i = a_{i, rl} + a_{i, cbf} + a_{i, clf}, \, i \in \{v,\theta\}$, we can rewrite the above fully actuated dynamics in the form in Equation \eqref{eq5D1}. Having obtained $[x_\epsilon, y_\epsilon]$, the robot's center coordinates can be calculated via a simple transformation.

\subsection{Task Definition}

The task is defined as

\begin{equation}\label{eq6C1}
    \begin{split}
        \phi = &(\diamondsuit \psi_{g_1} \vee \diamondsuit \psi_{g_2}) \wedge \diamondsuit \psi_{g_3} \wedge \\
        & (\neg \psi_{g_3} \,\, \mathcal{U} \,\,  (\psi_{g_1} \vee \psi_{g_2})) \wedge \\
         &((\psi_{c_1} \wedge \psi_{c_2}) \,\, \mathcal{U} \,\, \psi_{g_3})
    \end{split}
\end{equation}

\noindent In English, $\phi$ entails that the agent is to \textit{eventually} visit $g_1$ \textit{or} $g_2$ \textit{and eventually} $g_3$ \textit{and} $g_3$ is not to be visited before $g_1$ \textit{or} $g_2$ \textit{and} constraints $c_1$ and $c_2$ are to be satisfied at all times before $g_3$ is visited. The predicates are defined as $\psi_{g_i} = |\boldsymbol{p}_a - \boldsymbol{p}_{g_i}| < \eta_g, \, i \in \{1,2,3\}$ where $\boldsymbol{p}_a=[x,y]$ is the coordinate of the agent, $\boldsymbol{p}_{g_i}$ is the center coordinate of goal region $g_i$. $\psi_{c_1} = -y + x + w > 0$, $\psi_{c_2} = y - x + w > 0$ define the safety constraints as channel of width $2w$ that the agent needs to navigate within.

We define an additional reward $r = \underset{i}{\min}( |\boldsymbol{p}_a - \boldsymbol{p}_{o_i}|), \, i \in \{1,2\}$ as the minimum distance between the agent and the moving obstacles. This reward is part of the objective in Equation \eqref{eq3A1}.

\subsection{Implementation Details}

For the RL component, we use proximal policy optimization (PPO) \cite{schulman2017proximal} as the learning algorithm. The policy is represented by a feed-forward neural network with 3 hidden layers of 300, 200, 100 ReLU units respectively. The value function is of the same architecture. Each episode has horizon $T=200$ steps and the positions of the goal regions are randomized between episodes (goals may initiate outside the safe channel). We collect a batch of 5 trajectories for each update iteration. During learning, an episode terminates only when the horizon is reached or the task is completed. The agent is allowed to travel outside the safety channel and collide with the moving obstacles during learning (to receive a penalty). However, this is not allowed during evaluation.

\subsection{Results and Discussion}

In this section, we study the effect of each component in our system in Figure~\ref{fig:4.1}on the learning progress. Specifically, we train the system in four configurations - only RL, RL with CBF, RL with CLF and RL with CBF and CLF.

\begin{figure}[tbh]
\begin{center}
\includegraphics[width=0.85\linewidth]{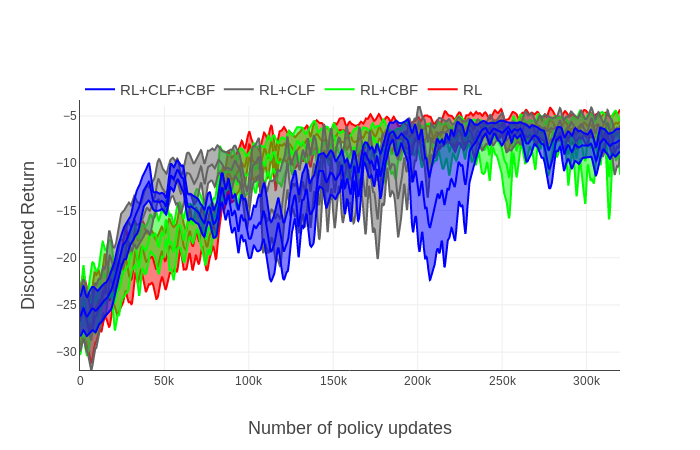}
\caption{Discounted return distribution for different experiment configurations. The figure reports the mean and 1 standard deviation calculated from a batch of 5 trajectories at each update iteration.}\label{fig:6.2.1}
\end{center}
\end{figure}

Figure \ref{fig:6.2.1}shows the learning curve in terms of the discounted return. We can see that all configurations are able to achieve similar end performance. The configurations with CLF exhibit faster rising time at the beginning of the learning process. This implies that the RL agent is able to find a useful policy faster with the help of CLF. However, it can also be observed that the learning curves with CLFs show higher variances and volatility. This is mainly due to the fact that the CLF is not aware of moving obstacles or safety constraints, its only job is to guide the agent to the nearest next goal that promotes progress on the FSA according to Equation \eqref{eq5B1}. Therefore, at moments of near collision and safety violations, the RL agent will need to fight the actions provided by the CLFs and hence adding difficulty to the learning problem. This is most apparent with the configuration RL+CLF+CBF where the RL agent will need to learn to work in harmony with the CLF and CBF controls.

\begin{figure}[tbh]
\begin{center}
\includegraphics[width=0.85\linewidth]{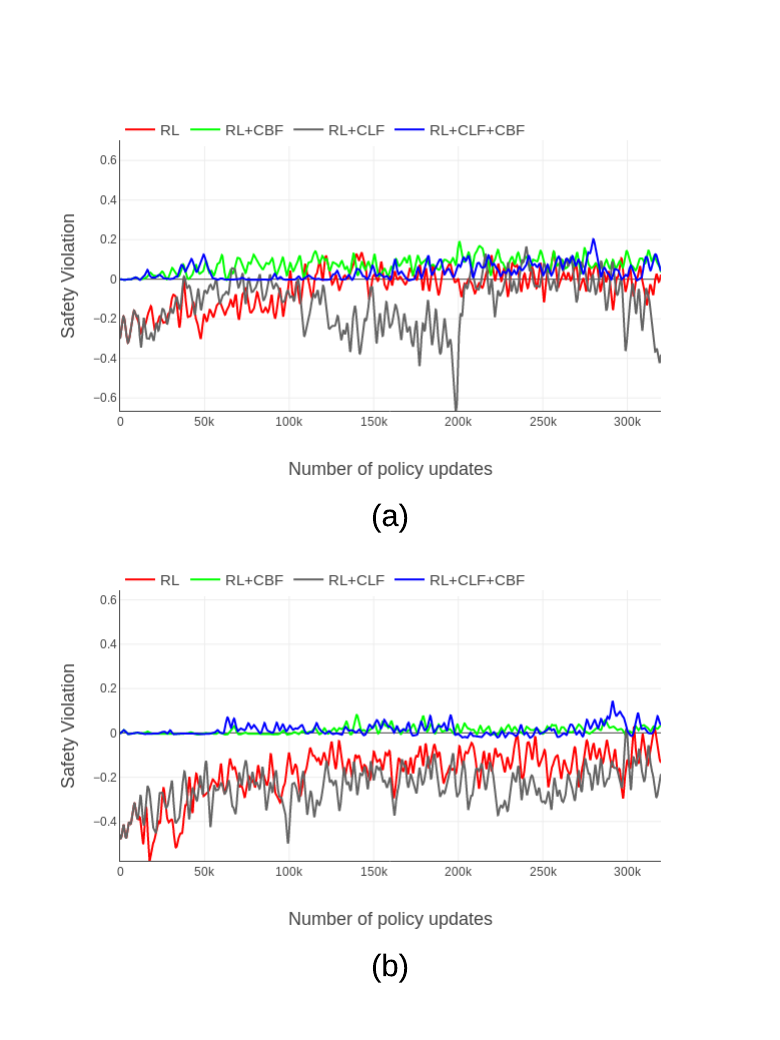}
\caption{Minimum value of the safety constraints for (\textbf{a}) $\psi_{c_1}$ and (\textbf{b}) $\psi_{c_2}$ calculated from a batch of 5 trajectories at each update iteration. A value smaller than 0 indicates violation of the constraints.}\label{fig:6.2.2}
\end{center}
\end{figure}

Figure \ref{fig:6.2.2}shows the maximum violation of the safety constraints $\psi_{c1}$ and $\psi_{c2}$ (a value smaller than zero indicates violation). We can see that the CBF is able to prevent the agent from violating the safety constraints at all times during learning. For configurations without CBF, because a violation penalty is imposed on the agent (Equation~\eqref{eq5A1}), it also gradually learns to minimize violation. Figure \ref{fig:6.2.2}(a) shows that the RL+CLF configuration exhibits a sudden decrease in performance (same time as in Figure~\ref{fig:6.2.1}). This occurs when the nearest next goal happens to be outside the safety region and the RL agent has not learned enough to counter the guiding actions of the CLF.

\begin{figure}[tbh]
\begin{center}
\includegraphics[width=0.85\linewidth]{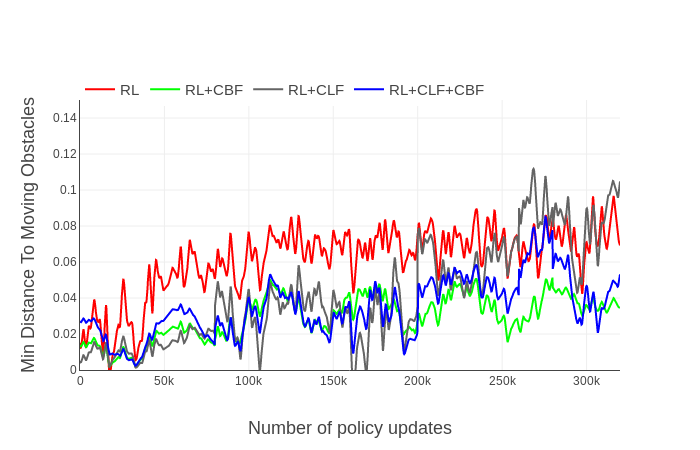}
\caption{Minimum agent distance to moving obstacles.}\label{fig:6.2.3}
\end{center}
\end{figure}

Figure \ref{fig:6.2.3} shows the minimum distance between the agent and the moving obstacles as a function of policy updates. As learning progresses, the agent learns to stay away from the obstacles. The agent is able to keep a further distance from the obstacles in configurations without CBF as this configuration allows the agent to travel outside the safe zone.

\begin{figure}[tbh]
\begin{center}
\includegraphics[width=0.9\linewidth]{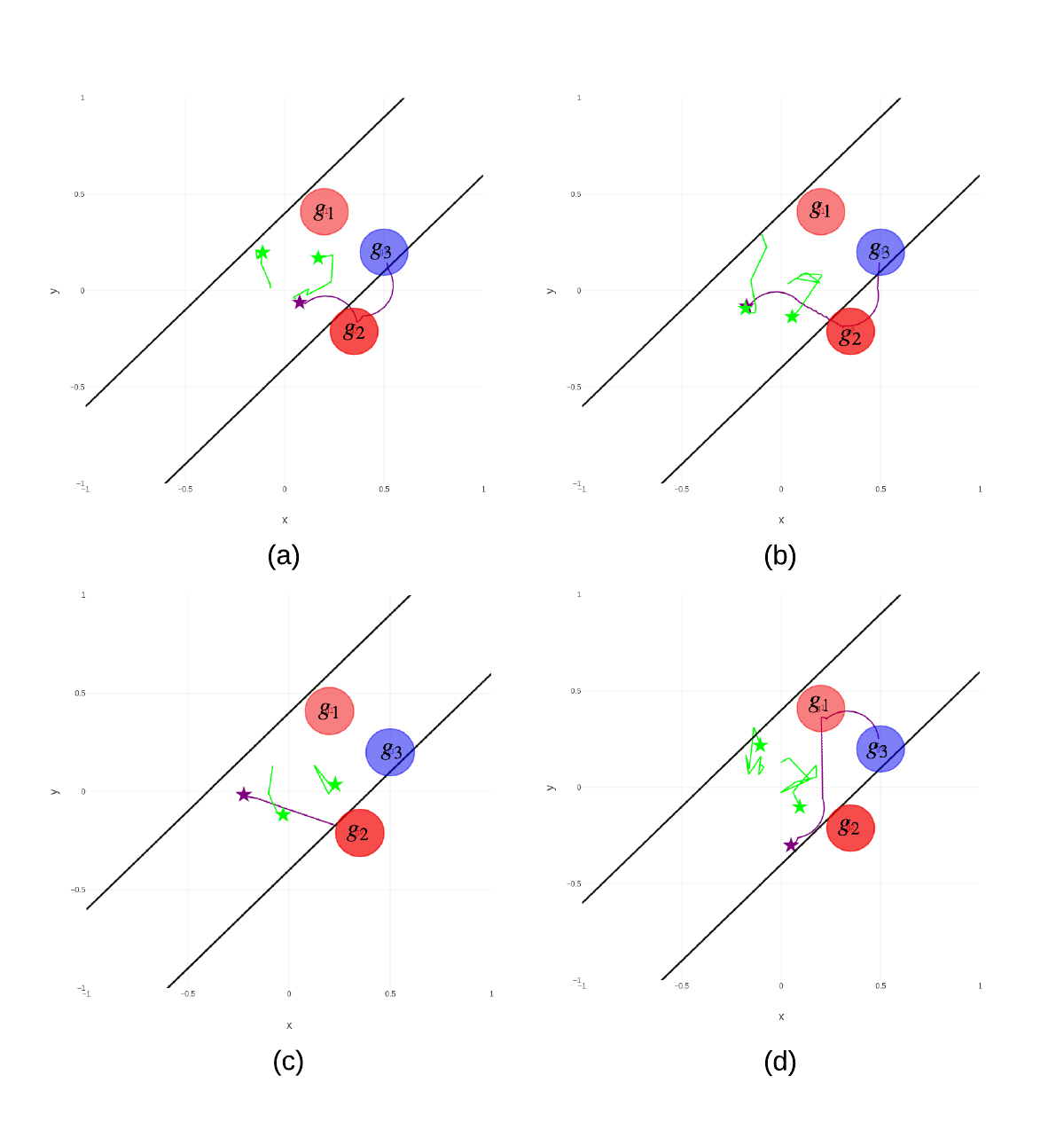}
\caption{Sample trajectories for agents with configuration (\textbf{a}) RL (\textbf{b}) CLF (\textbf{c}) CLF with CBF (\textbf{d}) RL, CLF and CBF. Agent trajectory is in purple, moving obstacle trajectories are in green. The stars represent starting positions.}\label{fig:6.2.5}
\end{center}
\end{figure}

Figure~\ref{fig:6.2.5} shows sample trajectories of the agent in various configurations. The upper sub-figures (a) and (b) illustrate trajectories for pure RL and pure CBF. We can see that  both agents behave similarly by going to $g_2$ (the nearer of $g_1$ and $g_2$) then $g_3$. The difference is that the RL agent chooses an obstacle free path and tries to trade off between accomplishing the task, avoiding obstacles and minimizing safety violation (controlled by the weights introduced in the previous section). Figure~\ref{fig:6.2.5}(c) shows the agent trajectory with CLF and CBF. CLF still commands the agent to $g_2$, however, CBF stops the agent at the border of the safety zone and the agent is stuck. Collision with moving obstacles occurs in this case because nor the CLF or the CBF know the obstacle dynamics which is required for avoidance. Figure~\ref{fig:6.2.5}(d) shows the agent trajectory with RL, CLF and CBF. The agent starts close to and tries to move towards $g_2$. However, it has learned that if it keeps trying to get to $g_2$ it will get stuck at the border and receive a low return. Therefore, near the border it chooses to instead move towards $g_1$ and eventually finishes the task. An agent trained with RL and CBF (not shown here) can potentially also obtain similar behavior. As discussed previously, adding CLF promotes exploration at the beginning of learning. A video can be accessed at https://youtu.be/lUsE3hGpLAk.

\begin{figure}[tbh]
\begin{center}
\includegraphics[width=0.8\linewidth]{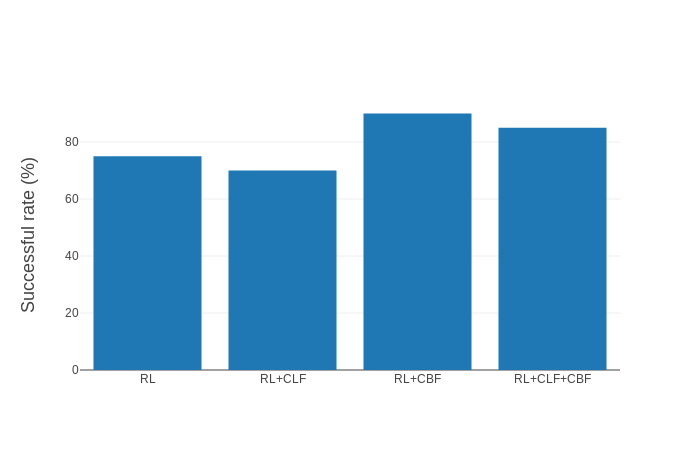}
\caption{Success rates over 20 evaluation trials.}\label{fig:6.2.4}
\end{center}
\end{figure}

Figure \ref{fig:6.2.4}shows the evaluation success rate over 20 trials. Different from the training process, during evaluation an episode terminates in three circumstances - horizon is reached, task accomplished and the agent comes in collision with the moving obstacles (defined by a minimum threshold on relative distance). To ensure safety, CBF is always turned on during evaluation. The results show that the agents trained with CBF exhibits higher success rates. This is because agents trained without CBF sometimes rely on traveling outside the safe zone to avoid obstacles and get to goals. During evaluation, this is not an option and therefore these agents are more prone to getting stuck at the safe zone boundaries.

Lastly, we want to discuss some practical issues of our approach and a possible area of future work. The system presented in Figure \ref{fig:4.1}integrates multiple components together. Central to controlling how the components interact with each other is a set of hyperparameters. This includes $c_r$ in Equation \eqref{eq5A1}, $w$ in Equation~\eqref{eq3A1}, the coefficients $K$, $\alpha$, $c_3$ and the control bounds in Equation \eqref{eq5D2}. We find that the learning process and the final performance of the policy is noticeably influenced by these parameters. Taking the control bounds as an example, a good practice is to set a wide bound for the CBF actions as these actions will need to dominate at the boundary of the safety constraints. The bounds for the CLF actions should be set narrower than that of the RL actions. This is to ensure that the RL agent is able to overcome the CLF actions when it needs to. A better strategy is to shrink the CLF action bounds as learning progresses such that it can guide exploration at the beginning of training and diminishes its effect as the agent learns. This type of guidance scheduling is also adopted in \cite{Rajeswaran2018LearningCD} with demonstrations. An important future work is therefore to alleviate the burden of parameter tuning from the user.

\section{CONCLUSIONS}

In this paper, we explore the combination of temporal logic, reinforcement learning, control Lyapunov functions and control barrier functions to form an expressive, safe and learnable control system. We propose techniques that allow an FSA to simultaneously provide rewards, objectives and safety constraints to each component in the system. We analyzed the effect of each component on the learning process and evaluate the resultant policy. In a simulation experiment with known agent dynamics and unknown environmental dynamics, we showed that our system is able to learn a policy that accomplishes the given task expressed as a logic formula while ensuring safety during learning.



\bibliographystyle{IEEEtran}
\bibliography{reference}

\end{document}